\documentclass[conference]{IEEEtran}
\IEEEoverridecommandlockouts

\usepackage{cite}
\usepackage{amsmath,amssymb,amsfonts}
\usepackage{algorithmic}
\usepackage{graphicx}
\usepackage{textcomp}
\def\BibTeX{{\rm B\kern-.05em{\sc i\kern-.025em b}\kern-.08em
    T\kern-.1667em\lower.7ex\hbox{E}\kern-.125emX}}

\usepackage{graphicx}
\usepackage{fancyhdr}
\usepackage{changepage} 
\usepackage{listings} 
\usepackage{url}

\usepackage[figurename=Fig., tablename=Tab., small]{caption}
\usepackage{graphicx}
\usepackage{amsmath}

\usepackage[table,xcdraw]{xcolor}
\usepackage{multirow}

\newcommand{\spf}{\mathbf{p}}
\newcommand{\spfset}{\MakeUppercase{{\spf}}}

\DeclareMathOperator{\countermeasure}{FDD}
\newcommand{\cm}[1]{\countermeasure(#1)}
\newcommand{\advflt}{\mathbf{m}}
\newcommand{\img}{\mathbf{I}}

\begin{document}

\title{2D-Malafide: Adversarial Attacks Against Face Deepfake Detection Systems
}

\author{\IEEEauthorblockN{1\textsuperscript{st} Chiara Galdi}
\IEEEauthorblockA{\textit{Digital Security Dep.} \\
\textit{EURECOM}\\
Sophia Antipolis, France}
\and
\IEEEauthorblockN{2\textsuperscript{nd} Michele Panariello}
\IEEEauthorblockA{\textit{Digital Security Dep.} \\
\textit{EURECOM}\\
Sophia Antipolis, France}
\and
\IEEEauthorblockN{3\textsuperscript{rd} Massimiliano Todisco}
\IEEEauthorblockA{\textit{Digital Security Dep.} \\
\textit{EURECOM}\\
Sophia Antipolis, France}
\and
\IEEEauthorblockN{4\textsuperscript{th} Nicholas Evans}
\IEEEauthorblockA{\textit{Digital Security Dep.} \\
\textit{EURECOM}\\
Sophia Antipolis, France}
}

\maketitle

\begin{abstract}
We introduce 2D-Malafide, a novel and lightweight adversarial attack designed to deceive face deepfake detection systems. Building upon the concept of 1D convolutional perturbations explored in the speech domain, our method leverages 2D convolutional filters to craft perturbations which significantly degrade the performance of state-of-the-art face deepfake detectors. Unlike traditional additive noise approaches, 2D-Malafide optimises a small number of filter coefficients to generate robust adversarial perturbations which are transferable across different face images.
Experiments, conducted using the FaceForensics++ dataset, demonstrate that 2D-Malafide substantially degrades detection performance in both \textit{white-box} and \textit{black-box} settings, with larger filter sizes having the greatest impact. 
Additionally, we report an explainability analysis using GradCAM which illustrates how 2D-Malafide misleads detection systems by altering the image areas used most for classification. Our findings highlight the vulnerability of current deepfake detection systems to convolutional adversarial attacks as well as the need for future work to enhance detection robustness through improved image fidelity constraints.
\end{abstract}


\section{Introduction}
In recent years, deep learning-based image recognition systems have achieved remarkable success across various applications, from face recognition to autonomous driving~\cite{Opanasenko_Fazilov_Mirzaev_Kakharov_2024,janai2020}. However, these systems are vulnerable to adversarial attacks, namely deliberate manipulations designed to deceive the model~\cite{fool_df_detect,9464957}.
Adversarial noise can typically be applied with subtle or seemingly insignificant perturbations to pixel values~\cite{adv_attacks_goodfellow}, involving even only small portions of the image. The perturbations are specially crafted to exploit model vulnerabilities and provoke erroneous outputs. Even if the perturbed image is indistinguishable to the eye from the original image, there can be drastic influences upon the model output.

Most adversarial attacks involve additive noise, where image-specific perturbations are learned and directly added~\cite{ambati2023prat}. Fortunately, these approaches are unsuitable for real-time implementation and exhibit high sensitivity to the specific input image. Typically, these methods are trained and tested using the same set of deepfake data, with no assurances of effectiveness against \emph{unseen} deepfakes — a property often referred to as \emph{generalisability}. 
Some adversarial attacks, whether additive~\cite{STANLY2023106595} or involving spatial transformations~\cite{9338327}, have partially solved the problem of generalisation but come at the cost of high complexity.

In this work, we propose the first adversarial attack which attempts to fulfil the \emph{generalisability} property through convolutive noise while still being computationally lightweight. The former goal is met by optimising the adversarial perturbation over multiple samples. The latter is achieved by reducing the number of learnable parameters thanks to simple, yet effective modelling choices.

Building on a previous work, named Malafide~\cite{malafide} which explored adversarial perturbation attacks against voice anti-spoofing solutions, we have tailored and implemented a novel adversarial attack named 2D-Malafide against image deepfake detection systems. This technique allows the attack to be mounted independently to the specific input image, and requires the optimisation of only a small number of filter coefficients. While the attack is agnostic to the type of classifier and image, e.g.\ be they face, fingerprint, or iris images, etc, in this paper we report its application specifically to face images and face deepfake detection.

Our experiments demonstrate that 2D-Malafide significantly degrades the performance of recent face deepfake detectors. The attack remains effective in both \textit{white-box} settings, where the filter is specifically trained to manipulate a particular detector model, as well as \textit{black-box} settings, and hence poses a substantial threat to the reliability of such detection systems.

\section{Related Work}

The concept of \emph{adversarial attacks} against neural networks was originally introduced in~\cite{intriguing_properties_of_neural_networks, adv_attacks_goodfellow} in the context of image classification tasks. The term usually refers to the introduction of perturbations to the input image of a neural network so as to manipulate the output or decision. Such perturbations can be crafted by optimising the pixel values of the input image via a gradient descent-based technique to maximise the output probability of an arbitrary, incorrect class.

Adversarial attacks have since been explored in a wide variety of different domains, including deepfake detection. Early investigations showed that deepfakes can be rendered undetectable by deepfake detection algorithms using specially crafted adversarial perturbations~\cite{carlini2020, fool_df_detect, hussain2021_adversarial, frequency_attacks}.
However, these studies focused on crafting individual adversarial perturbations for each deepfake sample, a computationally intensive process.

More recent adversarial attack techniques have since been proposed to overcome this issue. The authors of~\cite{fan2024} proposed the use of generative adversarial networks (GANs)~\cite{gans} to produce adversarial attacks for arbitrary deepfake samples.
In~\cite{Hou_2023_CVPR}, adversarial perturbations are modelled as a linear combination of image transformations whose weights are optimised across multiple deepfake images in order to minimise the chances of detection.
Using a similar objective function, the work in~\cite{Neekhara_2021_CVPR} demonstrates how a video deepfake detection system can be manipulated by using a single layer of additive noise with bounded amplitude applied to each image frame.

To the best of our knowledge, the only work that explores the generation of \emph{generalisable} adversarial perturbations against deepfake detectors
is~\cite{adv_moonlight_shadow}. The authors propose a GAN-based technique to produce shadows which are introduced to an image deepfake to conceal generated artefacts. Nonetheless, this technique involves the training of two generative neural networks and requires considerable computing capabilities.

\begin{figure*}[!t]
    \centering
    \includegraphics[width=\textwidth]{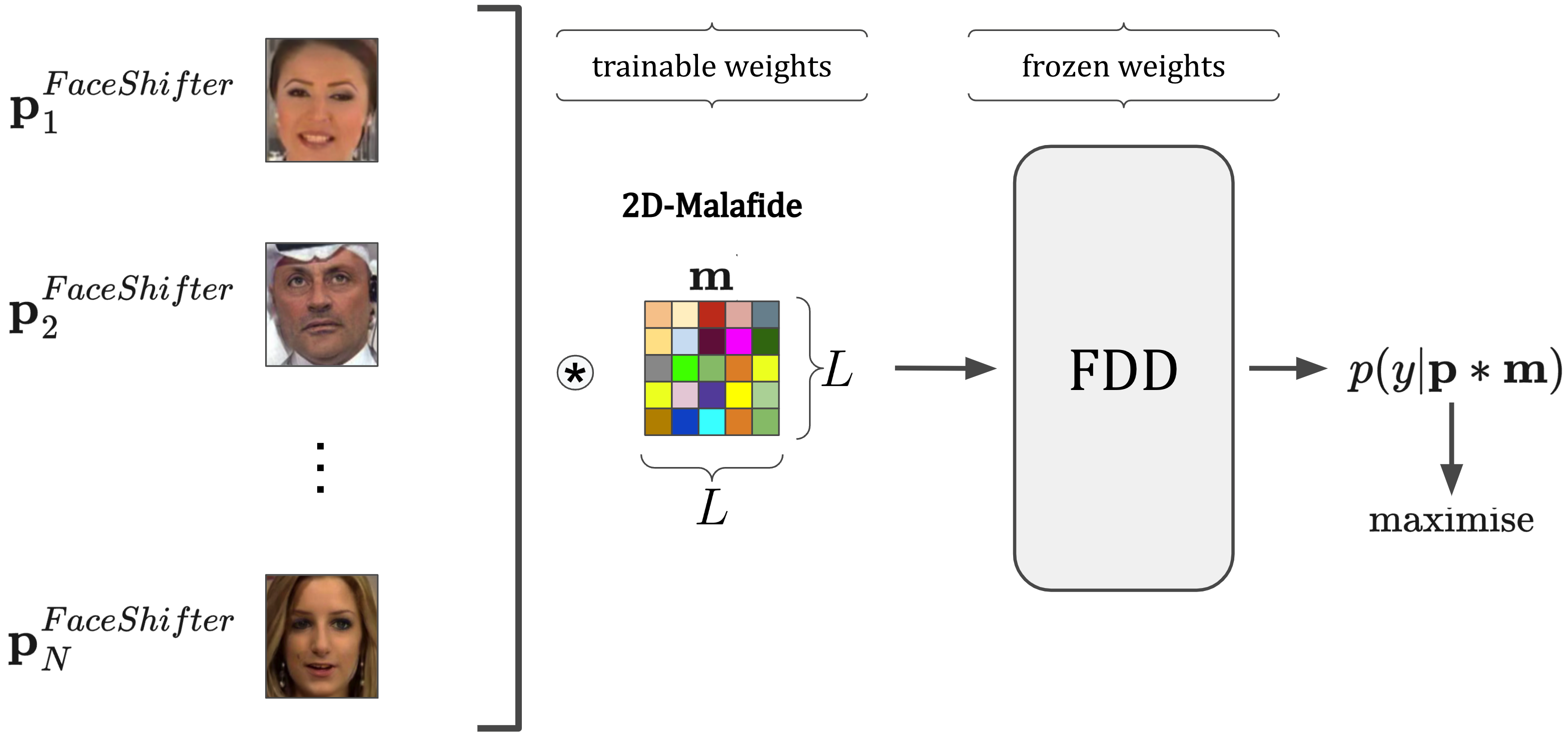}
    \caption{The training procedure of the 2D-Malafide filter $\advflt$  for face images generated with the attack $a=$ FaceShifter against the face deepfake detector (FDD).}
    \label{fig:training}
\end{figure*}

\section{2D-Malafide}

In this section we describe the adaptation and implementation of 2D-Malafide for adversarial attacks against face deepfake detection (FDD) systems.

Let $\spfset^{(a)} = \{\spf_1^{(a)}, \spf_2^{(a)} \dots \spf_N^{(a)}\}$ be a set of deepfake/spoofed images generated by algorithm~$a$. Each image is designed to deceive a deepfake detection system to increase the likelihood of false accept decisions. 
Let $\cm{\img} = s\left(y\mid\img\right)$ be a deepfake detector model which assigns a score $y$ to image $\img$, where higher scores reflect greater support for the bona fide class and lower scores for the deepfake class. For  spoofed images~$\spf_i^{(a)}$, $\cm{\spf_i^{(a)}}$ should hence produce low scores.
2D-Malafide attacks involve the optimisation of a 2D linear time-invariant (LTI), non-causal filter. The coefficients are optimised to provoke the misclassification of deepfake images as bona fide. The 2D LTI, $L \times L$ filter $\advflt^{(a)}$ is designed to maximise $\cm{\spf_i^{(a)} * \advflt^{(a)}}$, where $*$ denotes the 2D convolution operator. 
In the case of several different deepfake algorithms $a_1 \dots a_K$, an attacker can optimise an equivalent number of filters $\advflt^{(a_1)} \dots \advflt^{(a_K)}$. The filter should then be tuned to counter the reliance of the FDD system upon attack-specific artefacts.
Filter coefficients $\advflt^{(a)}$ can be optimised with conventional gradient descent using the set of spoofed images $\spfset^{(a)}$. The objective function is given by
\begin{equation}
    \label{eq:objective}
    \max_{\advflt^{(a)}} \sum_i \cm{\spf_i^{(a)} * \advflt^{(a)}}
\end{equation}
A graphical depiction of the training procedure is shown in Fig.~\ref{fig:training} for an attack $a=$ FaceShifter. The filter is optimised independently for each attack so as to manipulate the behaviour of a common FDD.

Without constraints, 2D-Malafide filtering can cause excessive image degradation. For detection settings in the absence of a human observer, this may have little consequence.  However, where the FDD system is deployed alongside other systems, the distortion introduced to compromise the FDD system might also interfere with the behaviour of any other auxiliary system, e.g.\ an automatic face recognition system. In this case, for instance, it might even \emph{improve} its resistance to attack, e.g.\ if image quality is significantly degraded.

Accordingly, $\advflt^{(a)}$ should be constrained to balance the maximisation of~\eqref{eq:objective} and the preservation of image fidelity, e.g., clarity, detail, or key features.
This can be achieved by tuning the filter size $L \times L$. Larger filters allow for greater manipulation and stronger attacks but can also introduce greater distortion. Conversely, smaller filters can be configured so that they introduce less distortion at the expense of a weaker attack.
We apply image normalisation after filtering in order to ensure that pixel values do not surpass the maximum quantisation level.

\section{Experimental Setup}

All experiments were conducted using the FaceForensics++ (FF++) dataset~\cite{roessler2019faceforensicspp}. It contains 1000 bona fide videos in addition to 5000 corresponding fakes generated with 5 different algorithms. 
The first two are computer graphics-based approaches. 
\textbf{Face2Face}~\cite{10.1145/3292039} is a facial reenactment system which transfers expressions from a source video to a target video while retaining the target face identity. 
\textbf{FaceSwap}~\cite{marek_marekkowalskifaceswap_2024} transfers the face region from a source to a target video using facial landmarks to fit a 3D model which is then backprojected, blended, and colour corrected. 
There are three deep learning-based approaches. The first, Deepfakes, was implemented using the open-source implementation \textit{deepfakes faceswap}\footnote{\url{https://github.com/deepfakes/faceswap}} and requires training with a pair of videos of source and target subjects. 
The second, \textbf{NeuralTextures}~\cite{10.1145/3306346.3323035}, learns a neural texture of the target person using a photometric reconstruction loss combined with an adversarial loss for training. The last is the two-stage face-swapping method \textbf{FaceShifter}~\cite{Li_2020_CVPR} which uses a pair of input images (a source for identity and a target for attributes like pose and expression) and a two-stage framework (AEINet and HEARNet) for high-fidelity face swaps. 

Although the FF++ dataset contains videos, the selected FDD systems operate on individual frames hence, in the remainder of this paper, mentions of the dataset refer to the collection of \textit{frames} extracted from FF++ videos. The attacker is assumed to have access only to the test partition of the dataset. Thus, the FF++ test partition was used for training and testing 2D-Malafide attacks. Attack-specific filters were trained according to (\ref{eq:objective}), using subsets of FF++ for each deepfake method. The FF++ test partition was split into 70\% for training (Part~1) and 30\% for testing (Part~2), with 1399 images in Part~1 and 599 images in Part~2. \mbox{2D-Malafide} filters were trained using Part~1 and tested using Part~2. This setup simulates offline filter training and online attacks. \mbox{2D-Malafide} filters were trained using only deepfake images.

Each attack-specific 2D-Malafide filter is trained using the Adam algorithm [24]. The learning rate and weight decay are tuned separately for each FDD system. The maximum number of epochs is set to 100 since, for all but a single experiment, training reaches the stop condition before 100 epochs, where the stop condition is defined by an equal error rate (EER) in excess of 50\%. The resulting filter is then applied to Part~2 for evaluation.  
A batch size of 32 was chosen because it was suitable for the GPUs used for our experiments. During optimisation of 2D-Malafide, the weights of the FDD pre-trained models are frozen. We explored different filter sizes $L=(3, 9, 27, 81)$ in order to analyse the impact on performance.
Our implementation is available as open-source and can be used to reproduce our results.\footnote{\url{https://github.com/eurecom-fscv/2D-Malafide}}

To determine the effectiveness of the adversarial filter attack we used the following two FDD systems.

\textbf{CADDM} \cite{Dong_2023_CVPR}\footnote{\url{https://github.com/megvii-research/CADDM}} is a deepfake detection system developed to address the problem of \textit{Implicit Identity Leakage}. The authors observed that deepfake detection models supervised using only binary labels are sensitive to identity.
Thus, they propose a method, termed an \textit{ID-unaware Deepfake Detection Model}, to reduce the influence of the identity representation. This is achieved by guiding the model to focus on local rather than global (whole image) features. Intuitively, by forcing the model to focus only on local areas of the image, less attention will be paid to global identity information.

\textbf{Self-Blended Images (SBIs)}~\cite{shiohara2022detecting}\footnote{\url{https://github.com/mapooon/SelfBlendedImages}} is a deepfake detection system which leverages training data augmentation to improve generalisability. The key idea behind SBIs is that the use of more general and barely recognisable fake samples encourage classifiers to learn generic and robust representations without overfitting to manipulation-specific artefacts. 
Fake samples are generated by blending pairs of \textit{pseudo source} and \textit{target} images, obtained using different image augmentation transformations, thereby increasing the difficulty of the face forgery detection task and encouraging the learning of more generalisable models. 

The implementations of both CADDM and SBIs used in this work support the use of different backbone architectures. For our experiments, both methods use EfficientNet convolutional neural networks, the only difference being that we use efficientnet-B3 for CADDM, but efficientnet-B4 for SBIs. Models pre-trained using the FF++ training dataset are used for both methods and are available on the respective GitHub repositories. 

\begin{table*}[!t]
\caption{Comparison in terms of EER [\%] of the baseline performance without filtering and the performance of different sizes of 2D-Malafide filters under \textit{white-box} (trained and tested on the same FDD) and \textit{black-box} (tested on different FDDs) settings. The results are shown for five attack types.}
\label{tab:PAD_combined}
\large
\centering
\begin{tabular}{ccccccccccc}
\hline
\rowcolor[HTML]{FFFFFF} 
                                  \multicolumn{11}{c}{\cellcolor[HTML]{FFFFFF}\textbf{Baseline Deepfake Detection System - CADDM (C) / SBI (S)}}                                                                                                                                                                                                                                                                                                                                                                        \\ \hline
\rowcolor[HTML]{FFFFFF} 
Attack type                           & \multicolumn{2}{c}{\cellcolor[HTML]{FFFFFF}Deepfakes}                                 & \multicolumn{2}{c}{\cellcolor[HTML]{FFFFFF}Face2Face}                                 & \multicolumn{2}{c}{\cellcolor[HTML]{FFFFFF}FaceShifter}                              & \multicolumn{2}{c}{\cellcolor[HTML]{FFFFFF}FaceSwap}                                  & \multicolumn{2}{c}{\cellcolor[HTML]{FFFFFF}NeuralTextures}                            \\ 
\hline

\cellcolor[HTML]{FFFFFF}FDD & \cellcolor[HTML]{FFFFFF}C              & \cellcolor[HTML]{FFFFFF}S              & \cellcolor[HTML]{FFFFFF}C              & \cellcolor[HTML]{FFFFFF}S              & \cellcolor[HTML]{FFFFFF}C             & \cellcolor[HTML]{FFFFFF}S              & \cellcolor[HTML]{FFFFFF}C              & \cellcolor[HTML]{FFFFFF}S              & \cellcolor[HTML]{FFFFFF}C              & \cellcolor[HTML]{FFFFFF}S              \\ \hline

\cellcolor[HTML]{FFFFFF}No filter & \cellcolor[HTML]{FFFFFF}0.00              & \cellcolor[HTML]{FFFFFF}0.71              & \cellcolor[HTML]{FFFFFF}1.34              & \cellcolor[HTML]{FFFFFF}1.43              & \cellcolor[HTML]{FFFFFF}1.34              & \cellcolor[HTML]{FFFFFF}7.14              & \cellcolor[HTML]{FFFFFF}0.67              & \cellcolor[HTML]{FFFFFF}1.43              & \cellcolor[HTML]{FFFFFF}2.50              & \cellcolor[HTML]{FFFFFF}5.00  

\\ \hline \\ \hline
\hline
\rowcolor[HTML]{FFFFFF} 
                                  \multicolumn{11}{c}{\cellcolor[HTML]{FFFFFF} \textbf{2D-Malafide trained on CADDM and tested on CADDM / SBI -  (W)hite box / (B)lack box}}                                                          \\
                                  \hline
\cellcolor[HTML]{FFFFFF} Filter size & W              & B              & W              & B              & W              & B              & W              & B              & W              & B              \\ \hline

\cellcolor[HTML]{FFFFFF}3x3       & \cellcolor[HTML]{D9EAD3}3.17              & \cellcolor[HTML]{D9EAD3}6.51             & \cellcolor[HTML]{D9EAD3}2.83              & \cellcolor[HTML]{D9EAD3}5.33             & \cellcolor[HTML]{D9EAD3}2.83              & \cellcolor[HTML]{F4CCCC}6.34             & \cellcolor[HTML]{D9EAD3}4.34              & \cellcolor[HTML]{D9EAD3}9.68             & \cellcolor[HTML]{D9EAD3}4.84              & \cellcolor[HTML]{D9EAD3}6.34            \\ \hline
\cellcolor[HTML]{FFFFFF}9x9       & \cellcolor[HTML]{D9EAD3}3.17              & \cellcolor[HTML]{D9EAD3}7.34              & \cellcolor[HTML]{D9EAD3}7.50              & \cellcolor[HTML]{D9EAD3}8.84              & \cellcolor[HTML]{D9EAD3}6.49              & \cellcolor[HTML]{F4CCCC}4.66              & \cellcolor[HTML]{D9EAD3}8.68              & \cellcolor[HTML]{D9EAD3}9.02              & \cellcolor[HTML]{D9EAD3}6.68              & \cellcolor[HTML]{D9EAD3}7.68             \\ \hline
\rowcolor[HTML]{F4CCCC} 
\cellcolor[HTML]{FFFFFF}27x27     & \cellcolor[HTML]{D9EAD3}46.41             & \cellcolor[HTML]{D9EAD3}8.01              & \cellcolor[HTML]{D9EAD3}49.83             & \cellcolor[HTML]{D9EAD3}7.16              & \cellcolor[HTML]{D9EAD3}50.17             & \cellcolor[HTML]{D9EAD3}7.16                                     & \cellcolor[HTML]{D9EAD3}46.41             & \cellcolor[HTML]{D9EAD3}7.68                                     & \cellcolor[HTML]{D9EAD3}51.92             & \cellcolor[HTML]{D9EAD3}6.01                                     \\ \hline
\rowcolor[HTML]{F4CCCC} 
\cellcolor[HTML]{FFFFFF}81x81     & \cellcolor[HTML]{D9EAD3}47.08             & \cellcolor[HTML]{D9EAD3}7.34              & \cellcolor[HTML]{D9EAD3}55.50             & \cellcolor[HTML]{D9EAD3}10.33              & \cellcolor[HTML]{D9EAD3}64.00             & 2.16                                     & \cellcolor[HTML]{D9EAD3}48.08             & \cellcolor[HTML]{D9EAD3}7.68                                     & \cellcolor[HTML]{D9EAD3}62.10             & 4.51                                     \\ \hline

\\ \hline
\hline
\rowcolor[HTML]{FFFFFF} 
                                  \multicolumn{11}{c}{\cellcolor[HTML]{FFFFFF} \textbf{2D-Malafide trained on SBI and tested on SBI / CADDM - (W)hite box / (B)lack box}}                                                          \\
                                  \hline
\cellcolor[HTML]{FFFFFF} Filter size & W              & B              & W              & B              & W              & B              & W              & B              & W              & B              \\ \hline

\cellcolor[HTML]{FFFFFF}3x3       & \cellcolor[HTML]{D9EAD3}6.18              & \cellcolor[HTML]{D9EAD3}3.17             & \cellcolor[HTML]{D9EAD3}13.17             & \cellcolor[HTML]{D9EAD3}2.83              & \cellcolor[HTML]{D9EAD3}11.00             & \cellcolor[HTML]{D9EAD3}2.83              & \cellcolor[HTML]{D9EAD3}8.01             & \cellcolor[HTML]{D9EAD3}3.67              & \cellcolor[HTML]{D9EAD3}13.86             & \cellcolor[HTML]{D9EAD3}4.51              \\ \hline
\cellcolor[HTML]{FFFFFF}9x9       & \cellcolor[HTML]{D9EAD3}13.86              & \cellcolor[HTML]{D9EAD3}1.34             & \cellcolor[HTML]{D9EAD3}28.83             & \cellcolor[HTML]{D9EAD3}1.50              & \cellcolor[HTML]{D9EAD3}34.17             & \cellcolor[HTML]{F4CCCC}0.67              & \cellcolor[HTML]{D9EAD3}31.39             & \cellcolor[HTML]{D9EAD3}2.00              & \cellcolor[HTML]{D9EAD3}33.06             & \cellcolor[HTML]{D9EAD3}2.84              \\ \hline
\rowcolor[HTML]{F4CCCC} 
\cellcolor[HTML]{FFFFFF}27x27     & \cellcolor[HTML]{D9EAD3}6.85              & \cellcolor[HTML]{D9EAD3}2.01             & \cellcolor[HTML]{D9EAD3}40.17             & \cellcolor[HTML]{D9EAD3}2.83              & \cellcolor[HTML]{D9EAD3}43.34             & \cellcolor[HTML]{F4CCCC}0.67                                      & \cellcolor[HTML]{D9EAD3}39.40             & \cellcolor[HTML]{D9EAD3}3.34                                      & \cellcolor[HTML]{D9EAD3}45.24             & \cellcolor[HTML]{D9EAD3}3.50                                      \\ \hline
\rowcolor[HTML]{F4CCCC} 
\cellcolor[HTML]{FFFFFF}81x81     & \cellcolor[HTML]{D9EAD3}29.05               & \cellcolor[HTML]{D9EAD3}3.17             & \cellcolor[HTML]{D9EAD3}26.67              & \cellcolor[HTML]{D9EAD3}2.83              & \cellcolor[HTML]{D9EAD3}45.99             & \cellcolor[HTML]{D9EAD3}2.83                                      & \cellcolor[HTML]{D9EAD3}30.05              & \cellcolor[HTML]{D9EAD3}5.01                                      & \cellcolor[HTML]{D9EAD3}29.22              & \cellcolor[HTML]{D9EAD3}3.84                                      \\ \hline

\end{tabular}
\end{table*}

\begin{figure*}[!t]
    \centering
    \includegraphics[width=\textwidth]{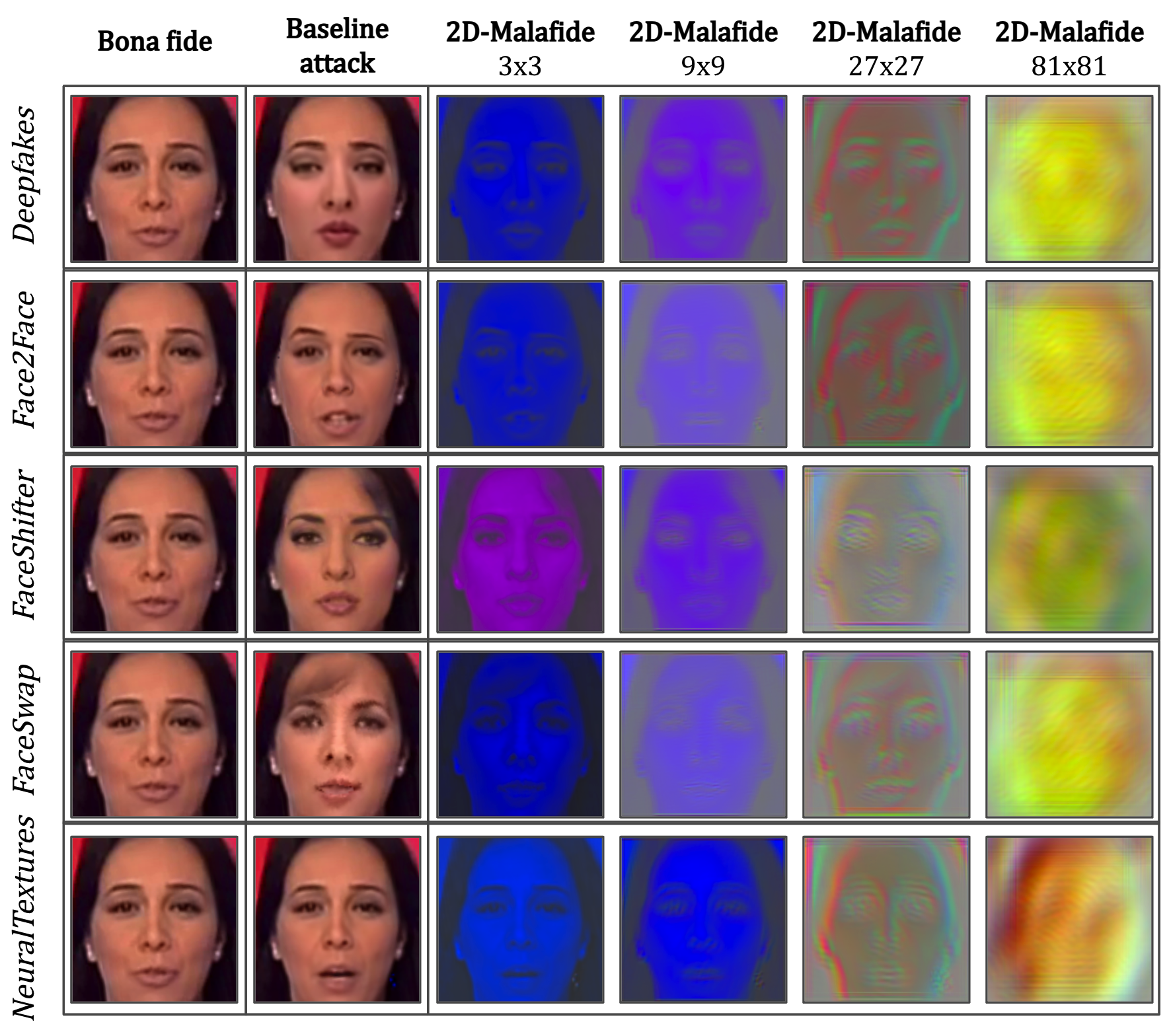}
    \caption{Examples of bona fide, baseline attack and four configurations of 2D-Malafide filter for the five deepfake attacks. Results are taken from training based on CADDM system.}
    \label{fig:examples}
\end{figure*}

\section{Experimental Results}\label{sec}

Results presented in Table \ref{tab:PAD_combined} show EER values for CADDM and SBI FDD systems with and without the application of 2D-Malafide filters under \textit{white-box} (tested using the same countermeasure used for 2D-Malafide training) and \textit{black-box} (tested using an unseen countermeasure) settings. 
Results are shown separately for the baseline FDD system (top block), then 2D-Malafide attacks trained using CADDM (middle) and SBI (bottom).
Baseline FDD results show detection error rates for the five different attack types.

For the CADDM \textit{white-box} setting (denoted W in Table~\ref{tab:PAD_combined}), the application of 2D-Malafide filters leads to a significant increase in EER, especially with larger filter sizes ($27\times27$ and $81\times81$). This indicates a substantial degradation in FDD performance, demonstrating the effectiveness of the adversarial filters in deceiving the detection system.
For the corresponding \textit{black-box} setting, for which the model is trained using CADDM but tested using SBI, results show that most filters provoke an increase in the baseline EER. However, in some cases (highlighted in red), filtering instead reduces the EER, indicating that they made it easier for the FDD system to detect the underlying attack.

For the SBI \textit{white-box} setting the 2D-Malafide filters again lead to notable increases in the EER, particularly for the $27\times27$ filter size. We note that, for the $81\times81$ filter, the EERs decrease slightly, showing that the largest filter size is less effective for SBI than for CADDM. For the corresponding \textit{black-box} setting, filtering generally increases the baseline EER. However, the impact is less pronounced compared to CADDM, indicating that adversarial training performed using SBI does not generalise well.

Overall, results indicate that FDD systems are vulnerable to 2D-Malafide attacks, with the greatest impact observed under \textit{white-box} settings. The impact varies with filter size. Larger filters ($27\times27$ and $81\times81$) tend to cause the most significant degradation in detection performance, particularly for CADDM. Under \textit{black-box} settings, while filtering generally provokes an increase in error rates, there are instances where detection performance improves, suggesting that adversarial filtering does not always generalise well to unseen detectors.

Last, Fig.~\ref{fig:examples} shows a comparison of bona fide images, the  corresponding attacks and then after application of four different 2D-Malafide filters, for the CADDM FDD system. For smaller filter sizes, the face is still recognisable even if the colours are unnatural.
For larger filters, the face is significantly distorted or even  unrecognisable. This finding in itself highlights a critical limitation in face deepfake detection in that they can be compromised so easily with images which do not even resemble natural faces. 
This raises concerns about the robustness of such systems when dealing with altered or degraded images, with obvious implications for both the security and reliability of the technology.

\begin{figure*}[!t]
    \centering
    \includegraphics[width=\textwidth]{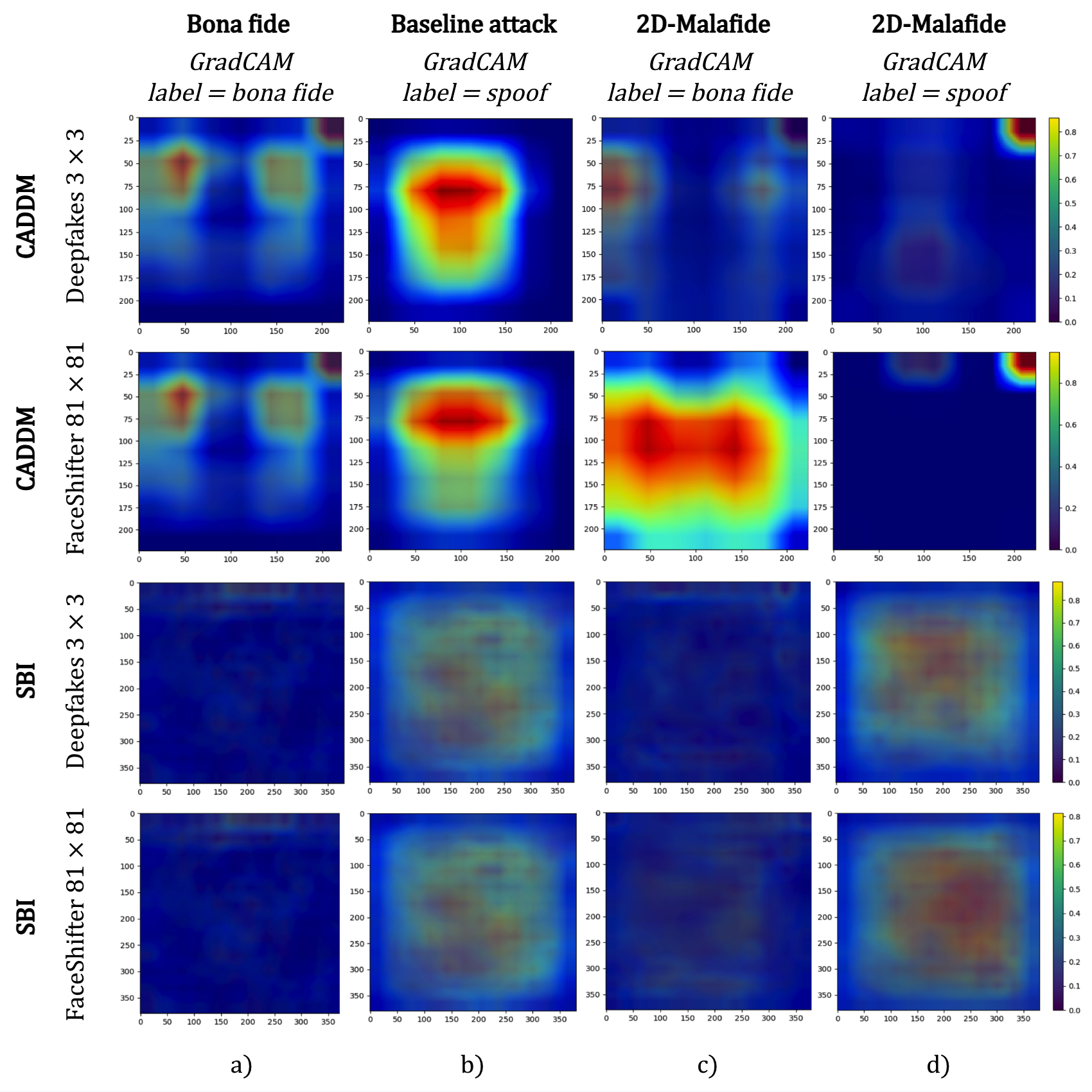}
    \caption{GradCAM explainability results for Deepfakes $3\times3$ and  FaceShifter $81\times81$ image samples classified with CADDM and SBI FDD systems applied on bona fide (a), baseline attack (b), and 2D-Malafide attacks processed with GradCAM label bona fide (c) and spoof (d).}
    \label{fig:explainability5}
\end{figure*}

\section{Explainability Analysis}

In order to gain deeper insights into the impact of 2D-Malafide filtering upon the deepfake detectors, we also report an explainability analysis. We report the GradCAM~\cite{jacobgilpytorchcam} heatmaps for a pair of different attacks and filter sizes when using the CADDM and SBI detectors, specifically Deepfakes $3\times3$ and FaceShifter $81\times81$.
GradCAM is applied to each image in the test set and for each category: \textit{bona fide}, \textit{spoof}, \textit{spoof + malafide}.  The resulting heatmaps are averaged to show predominant activation patterns.

The heatmaps in Fig.~\ref{fig:explainability5} indicate the areas of the face images where the model focuses its attention according to the input label, hence revealing features relevant to either \textit{bona fide} or \textit{spoof} classes.
The first row of Fig.~\ref{fig:explainability5} shows  results for CADDM and the Deepfakes $3\times3$ attack. The left-most heatmap in column (a), shows that significant facial landmarks which correspond to the contours of the face are most informative for the classification of images as bona fide. In constrast, in the case of fake images, shown in column (b), facial landmarks corresponding to areas of the eyes and eyebrows are most informative. Visual inspections reveal that these areas often correspond to visible artefacts, e.g.\ double eyebrows, resulting from the application of Deepfakes.

Heatmaps in columns (c) and (d) display results after application of 2D-Malafide and for fake face images when using \textit{bona fide} and \textit{spoof} labels respectively. Whereas heatmaps (a) and (c) exhibit similar patterns, heatmaps (b) and (d) are notably different. 2D-Malafide hides fake image artefacts upon which the detector relies, namely those in the central part of the face. There are no obvious activations in this area in heatmap (d), hence why the model is misled into classifying the fake as bona fide.

The second row of Fig.~\ref{fig:explainability5} shows results for FaceShifter $81\times81$ attacks, again for CADDM. The heatmap in column (c) shows that the CADDM model focuses on the sides of the face image, but with greater intensity than for bona fide images. Heatmap (d) remains similar to that for the Deepfakes $3\times3$ attack. Not only does 2D-Malafide hide fake artefacts, it also provokes a greater rate in the misclassification of fake images by causing the detector to focus more on sides of the face. This finding accounts for results reported in Table~\ref{tab:PAD_combined}, in particular cases for which 2D-Malafide is more efficient the largest filter size.
The dominant spot to the upper right might be due to the Multi-scale Detection Module (MSDM) of the CADDM architecture. The MSDM uses predefined anchor boxes which are tiled across the image. The level of activations in this area might correspond to the location of the last analysed anchor box.

Heatmaps in rows 3 and 4 of Fig.~\ref{fig:explainability5} show results for the SBI detector. For the Deepfakes $3\times3$ attack, the detector focuses on small parts of bona fide images at different positions, hence the seemingly flat heatmap. In contrast and in the case of fakes, the model focuses predominantly on central areas of the face, albeit in a less localised manner compared to CADDM. After application of 2D-Malafide filtering, there are few differences between results for bona fide images~(a) and filtered bona fide images~(c), and also between those for fakes~(b) and filtered fakes~(d). However, a closer look revels how attention for attacks without filtering, shown in column (b), is more concentrated to the bottom left of the central part of the face. Instead, for fake images processed by 2D-Malafide, attention is concentrated more to the top right, and more so for FaceShifter $81\times81$ attacks.

\section{Conclusions}

In this article we introduce 2D-Malafide, an adversarial attack which uses 2D convolutional filtering to deceive face deepfake detection systems. The attack significantly increases the EER of state-of-the-art deepfake detectors in both \textit{white-box} and \textit{black-box} settings and highlights the vulnerability of current FDD systems to such attacks. Larger filters ($27\times27$ and $81\times81$) cause substantial performance degradation. Moreover, the generalisability of 2D-Malafide ensures robustness across various image inputs, making for a versatile threat.
Colour information is the first to be impacted by the application of 2D-Malafide showing that the FFDs considered in this work fail to recognise simple, even unnatural changes in colour.

GradCAM explainability analysis reveals that 2D-Malafide misleads FDD systems by altering the areas of an image they use for classification, thereby increasing false acceptance rates. Attack success varies across different FDD systems, indicating some level of generalisability but also a dependency on the specific architecture.

The results emphasise the need for comprehensive and diverse training datasets to improve FDD robustness. Future research should focus on enhanced image fidelity constraints, including colour consistency, to counter such adversarial attacks. Overall, 2D-Malafide demonstrates the critical need for ongoing advancements in FDD technology to ensure the security and reliability of deepfake detection systems.

\bibliographystyle{ieeetr}
\bibliography{biblio}

\end{document}